\documentclass[10pt,twocolumn,letterpaper]{article}

 \usepackage{iccv}              

\usepackage{comment}
\usepackage{array}
\usepackage{tabularx}
\usepackage{graphicx}
\usepackage{subcaption}
\usepackage{float}
\usepackage{amsmath}
\usepackage{placeins}
\usepackage{caption}

\usepackage{algorithm}
\usepackage{algpseudocode}  

\definecolor{iccvblue}{rgb}{0.21,0.49,0.74}
\usepackage[pagebackref,breaklinks,colorlinks,allcolors=iccvblue]{hyperref}

\def\paperID{*****} 
\def\confName{ICCV}
\def\confYear{2025}

\title{ProtoMedX: Towards Explainable Multi-Modal Prototype Learning for Bone Health Classification}

\author{%
  Alvaro Lopez Pellicer\textsuperscript{1} \quad André Mariucci\textsuperscript{1} \quad Plamen Angelov\textsuperscript{1}\\
  Marwan Bukhari\textsuperscript{2} \quad Jemma G. Kerns\textsuperscript{2}\\
  {\small \textsuperscript{1}School of Computing and Communications; \textsuperscript{2}Lancaster Medical School}\\
  {\small Lancaster University, Lancaster, United Kingdom}\\
  {\ttfamily\small \{a.lopezpellicer,a.mariucci,p.angelov,m.bukhari,j.kerns\}@lancaster.ac.uk}
}

\begin{document}
\maketitle
\begin{abstract}
Bone health studies are crucial in medical practice for the early detection and treatment of Osteopenia and Osteoporosis. Clinicians usually make a diagnosis based on densitometry (DEXA scans) and other patient history. The applications of AI in this field are an ongoing research. Most of the successful methods for this task include Deep Learning models that rely on vision alone (DEXA / X-ray imagery) geared towards high prediction accuracy, where explainability is disregarded and largely based on the \textit{post hoc} assessment of input contributions. We propose \textbf{ProtoMedX}, a multi-modal model that uses both DEXA scans of the lumbar spine and patient records. ProtoMedX's prototype-based architecture is explainable by design, crucial for medical applications, especially in the context of the upcoming EU AI Act, as it allows explicit analysis of the model's decisions, especially the ones that are incorrect. ProtoMedX demonstrates state-of-the-art performance in bone health classification while also providing explanations that can be visually understood by clinicians. Using our dataset of 4,160 real NHS patients, the proposed ProtoMedX achieves 87.58\% accuracy in vision-only tasks and 89.8\% in its multi-modal variant, both approaches surpassing existing published methods.
\end{abstract}    
\section{Introduction}
In 2022, osteoporosis affected $3.5$ million people in the United Kingdom, with osteoporotic fractures accounting for over £$4.6$ billion of direct costs to the National Health Service (NHS). This annual cost is forecasted to rise to £$6$ billion by 2030 as the population of the United Kingdom ages \cite{ros2022evidence}. Globally, the prevalence of osteoporosis is estimated to be $6.3\%$ of men over the age of 50 and $21.2\%$ of women over the same age range \cite{iof_epidemiology_fragility_fractures_2025}. Based on the world population, this suggests that approximately $500$ million individuals worldwide may be affected~\cite{iof_epidemiology_fragility_fractures_2025}. 


Existing bone health classification approaches face three fundamental limitations. First, most studies adopt binary classification (normal vs. osteoporosis), either excluding osteopenia patients altogether or grouping them with osteoporosis cases under the label ``low bone density" \cite{9752533}. This simplification obscures the clinically important intermediate state of osteopenia, where patients face a significantly higher fracture risk compared to those with osteoporosis \cite{TomasevicTodorovic2018}. Possibly due to data availability issues, studies attempting three-class classification (normal, osteopenia, and osteoporosis) are much rarer. The only three-class study we identified using spine data, \cite{ZHANG2020115561}, did not report classification accuracy but achieved a highest AUC of 0.81, suggesting that the added complexity poses substantial challenges.

Second, current methods predominantly rely on vision-only models \cite{sukegawa2022identification, luan2025application}. This contradicts standard medical practice for fracture risk assessment, where diagnosis integrates image findings with patient history, demographics, and risk factors. Our results validate this intuition; by incorporating clinical features alongside imaging, accuracy improves from \textbf{87.58\% to 89.8\%} while providing more robust predictions.


Third, and critically, existing deep learning approaches lack explainability. \textit{Post hoc} interpretation methods such as GradCAM \cite{selvaraju2019grad} and SHAP values \cite{NIPS2017_7062} provide approximate rationalisations post-training rather than true model reasoning. In medical contexts where decisions directly impact patient care, clinicians require models that are explainable by design and cannot rely on ``black boxes" with retrofitted interpretations to guide their treatment decisions \cite{rudin2019stop}. This importance is further emphasised in the context of the recently adopted EU AI Act, which considers medical applications of AI as ``high risk", making it as significant as critical infrastructure ~\cite{onitiu2023limits, europeanunion2024aiact, europeanunion2024aiact}.

To address these limitations, we present \textbf{ProtoMedX} (\textbf{Proto}type-based \textbf{Med}ical e\textbf{X}planation), the first prototype-learning framework for bone health assessment. ProtoMedX reimagines bone health classification through case-based reasoning, rather than learning opaque decision boundaries. It identifies representative prototypes for each diagnostic category and classifies new patients based on similarities to these learned examples. This mirrors actual clinical reasoning, where physicians compare patients to archetypal cases. Our key contributions are:

\begin{itemize}
\item \textbf{First prototype-based architecture for bone health}: We introduce dual prototype spaces (visual and clinical) unified through cross-modal attention, enabling explainable predictions via direct comparison to learned exemplars. (See Figures~\ref{fig:protomedx_architecture},~\ref{fig:protomedx_horizontal}). 

\item \textbf{Multi-task learning leveraging bone density continuity}: We demonstrate that auxiliary T-score regression during training forces the model to understand bone density as a continuous phenomenon, dramatically improving classification accuracy by 6.7\% in MLP approaches (Table~\ref{tab:performance}) and by 2.46\% in ProtoMedX (See Table~\ref{tab:ablation}).

\item \textbf{State-of-the-art performance with built-in explainability}: ProtoMedX achieves 89.8\% accuracy on three-class classification, surpassing prior art by 14-27\% absolute improvement (Table \ref{tab:international}). Critically, this performance includes inherent explanations, not \textbf{post hoc} rationalisations, making it more suitable for clinical deployment.

\item \textbf{Comprehensive evaluation on clinical data}: The study is comprised of 4,160 lumbar spine DEXA scans with complete patient profiles, demonstrating robust performance across demographics. The model excels at clinically critical Normal vs. Abnormal detection (91.2\% sensitivity) while maintaining prototype-based transparency. (See Sec.~\ref{subsec:dataset}).
\end{itemize}
\section{Background and Related Work}

\subsection{Bone Health}

As we age, we naturally loose bone mass. This happens to some patients at a faster rate than others, which can result in osteoporosis. Osteoporosis is a bone disease that results in an increased risk of a fragility fracture, which occurs following a fall from standing height or less. Osteoporosis affects both men and women but is most prevalent in postmenopausal women due to reduced oestrogen, with 1 in 2 women affected by the age of 80 \cite{iof_epidemiology_fragility_fractures_2025}. While it is not currently possible to completely recover bone after significant degradation, it is possible to reduce further bone loss and the impact through lifestyle and/or pharmaceutical interventions if detected sufficiently early. There are several clinical tools available to assess fracture risk, most notably FRAX® \cite{kanis2025frax}, which calculates a patient's probability of suffering from a hip fracture caused by osteoporosis within 10 years, as well as provides advice as to whether a patient should seek a bone densitometry scan produced by a DEXA machine \cite{nogg2024manual}. The DEXA machine (Dual-Energy X-ray Absorptiometry) uses a low dose of X-rays to measure bone mineral density and provides a T-score. The T-score indicates the number of standard deviations away a patient is from a young, healthy population and provides the diagnostic definition of bone health, according to the World Health Organization (WHO), using the following thresholds \cite{ros2025dxa}: normal (T-score above $-1.0$), osteopenia (T-score between $-1.0$ and $-2.5$), or osteoporosis (T-score below $-2.5$) \cite{ros2025dxa}.

DEXA scans are commonly performed on hips, forearms, lumbar spine, and whole-body scans. For this study, we were granted the use of lumbar spine scans only.

\subsection{Dataset Availability}

Moderate-sized tabular datasets (1,000–10,000 samples) are available, such as~\cite{luan2025application} who employed a sample of 6,672 patients from an NHANES-provided dataset \cite{NHANES_CDC}. For vision tasks, publicly available datasets for use in research are often significantly smaller, normally in the hundreds, which limits the amount of diversity for each dataset and limits the quality of research by third-party institutions. One example of a publicly available dataset includes 177 lumbar DEXA scans from patients in Pakistan \cite{malik2021osteoporosis}. These public datasets are usually limited in size due to strict data restrictions, often resulting in datasets remaining proprietary. To the best of our knowledge, there are no large or moderate sized datasets that contain both images and more than a few patient attributes (outside of age, weight, height, and gender) that are available to the public, making multi-modal machine learning largely inaccessible. While our dataset is not currently accessible to the public, we support ongoing efforts to promote the sharing and accessibility of such datasets within the research community, provided the necessary data protection procedures are correctly implemented. Due to the aforementioned limitations in data availability, many studies~\cite{9752533,ZHANG2020115561,lee2019osteoporosis} rely on data augmentation methods to resolve imbalanced data, which is standard for vision tasks. However, there are doubts on the transferability to real clinical practice application of models that leverage augmentation within the medical domain \cite{pattilachan2023critical,tupper2024analyzing,pohjonen2022augment}.

Our research was performed in collaboration with the NHS, which allowed us to obtain a significantly larger dataset than is common in this field. The dataset contains 4,160 patients' lumbar spine DEXA scans, of which 2,224 are patients with normal bone health, 1,398 are patients with osteopenia, and 538 are patients with osteoporosis. The data consists of 926 males and 3,234 females. The ages of patients in the dataset range from 20 to 110, with an average age of 74 and an interquartile range of 65 to 83. The dataset also contains parts of the patients records, specifically, the same 11 features that the FRAX® tool operates with.


\subsection{Machine Learning for Bone Health}

Machine learning approaches to bone health classification have evolved from early neural networks~\cite{Ongphiphadhanakul1997BMD_ANN, Queralto1999BoneLossANN} to modern deep learning architectures. However, fundamental limitations persist across these methods. \\

\noindent\textbf{Tabular-only approaches}, such as those using demographic and bone turnover markers (e.g., age, sex, and bone-specific alkaline phosphatase), can achieve up to 73\% accuracy~\cite{Baik2024OsteoporosisML}. However, these methods lack the spatial information provided by imaging. \\


\noindent\textbf{Vision-only methods} using CNNs and Vision Transformers on DEXA/X-ray images report 80-90\% accuracy~\cite{luan2025application, sukegawa2022identification, sarmadi2024comparative, 9752533}. However, these studies suffer from (1) binary classification that ignores the clinically important osteopenia stage~\cite{9752533}, (2) small datasets (e.g., 117 images~\cite{9752533}) with heavy augmentation, (3) uncommon body areas, such as~\cite{sarmadi2024comparative} using knees, which is not commonly used for osteoporosis diagnosis, or (4) poor three-class performance \cite{ZHANG2020115561}.\\

\noindent\textbf{Multi-modal fusion} shows modest improvements. \cite{sukegawa2022identification} achieved 84.5\% (vs. 83.2\% vision-only) by combining dental X-rays with demographics. While this study employed binary classification, it demonstrates that the addition of tabular features in a multi-modal setting can provide increased accuracy, which is in line with our own findings.

These approaches share critical gaps: oversimplified binary classification, lack of explainability, and limited multi-modal integration, motivating our prototype-based solution that addresses all three.

\subsection{Explainable AI for Bone Health}

Clinical deployment demands models that are able to explain their reasoning, not just achieve high accuracy. \textit{Post hoc} methods such as GradCAM \cite{selvaraju2019grad}, SHAP \cite{NIPS2017_7062}, and saliency maps \cite{zeiler2014visualizing} approximate model behaviour after training but lack fidelity and have the capability of being misleading, as they provide rationalisations rather than true model reasoning \cite{rudin2019stop,hooshyar2024problems,krishna2022disagreement,witzel2024explainable,gosiewska2020donottrust}. This disconnect is particularly problematic in medical contexts where understanding the decision process guides treatment.

\begin{figure}[ht]
    \centering
    \begin{subfigure}[b]{0.3\columnwidth}
        \centering
        \includegraphics[width=\linewidth,clip,trim=0 100 0 0]{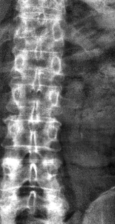}     
        \vspace{0.5em}
        \parbox{\linewidth}{\centering \small \textbf{1a.} Good prototype.}
    \end{subfigure}
    \hspace{5em}
    \begin{subfigure}[b]{0.3\columnwidth}
        \centering
        \includegraphics[width=\linewidth]{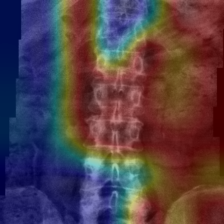} 
        \vspace{0.5em}
        \parbox{\linewidth}{\centering \small \textbf{1b.} Poor heatmap localisation.}
    \end{subfigure}
    \vspace{1em}
    \begin{subfigure}[b]{0.3\columnwidth}
        \centering
        \includegraphics[width=\linewidth]{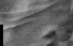}       
        \vspace{0.5em}
        \parbox{\linewidth}{\centering \small \textbf{2a.} Poor prototype.}
    \end{subfigure}
    \hspace{5em}
    \begin{subfigure}[b]{0.3\columnwidth}
        \centering
        \includegraphics[width=\linewidth]{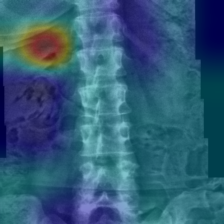}        
        \vspace{0.5em}
        \parbox{\linewidth}{\centering \small \textbf{2b.} Good heatmap localisation.}
    \end{subfigure}
    \caption{Inconsistent quality of prototypes and heatmap localisation generated by ProtoPNet. Despite (1a) showing a clear, high-quality prototype, its heatmap (1b) exhibits poor localisation with diffuse activation. Conversely, a blurry prototype (2a) produces well-focused heatmap localisation (2b), revealing that prototype visual quality does not correlate with localisation accuracy.}
    \label{fig:full_grid}
\end{figure}

\paragraph{Prototype-based Models:} Prototype learning offers inherent explainability by comparing inputs to learned exemplars. ProtoPNet \cite{Chen2019} pioneered this approach for images, followed shortly by xDNN \cite{Angelov2020xDNN}, enabling case-based reasoning (``this looks like that"). However, ProtoPNet suffers from three limitations: (1) vision-only architecture makes the framework incompatible with clinical data, (2) reliance on localised patches that can highlight misleading regions, and (3) \textit{post hoc} similarity heatmaps rather than supervised prototype learning \cite{tegerdine_willard_triplett_tang_2023} (see Figure~\ref{fig:full_grid}).

To the best of our knowledge, no prior work applies prototype-based learning to bone health classification. \textbf{ProtoMedX} addresses this gap by extending prototype reasoning to multi-modal inputs (imaging + clinical data) while learning full-image prototypes that provide clear explanations. This design aligns with the way clinicians reason, comparing patients to archetypal cases rather than analysing isolated image regions.
\section{ProtoMedX}
\begin{figure*}[t]
    \centering
    \includegraphics[width=\textwidth]{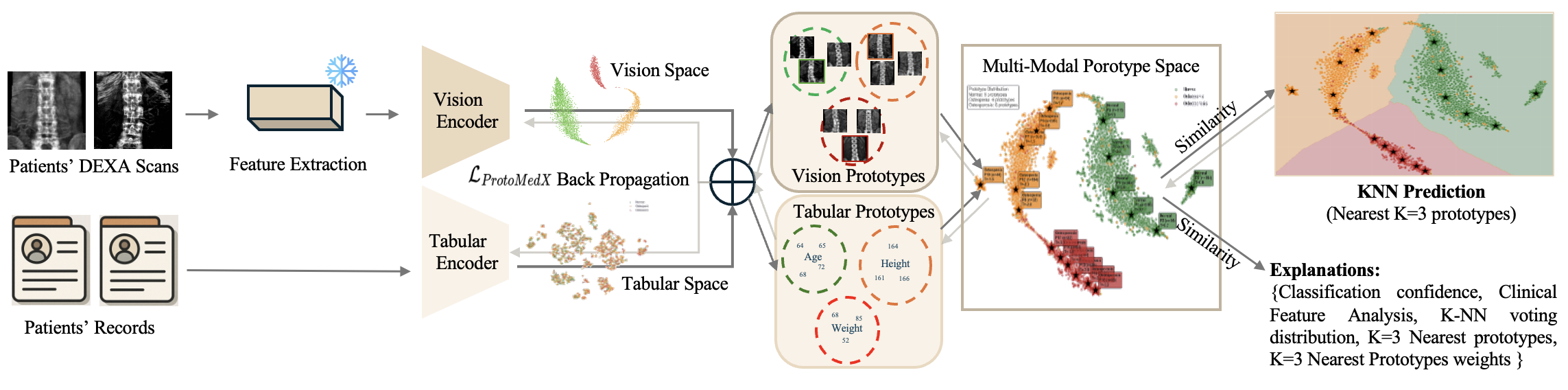}
    \caption{
        \textbf{Overview of ProtoMedX Architecture.} 
        Multi-modal prototype learning combines patient DEXA scans and clinical records \textit{via} separate encoders, learns explainable vision and tabular prototypes, and fuses them in a joint prototype space. Classification and explanations derive from prototype similarity and case retrieval, enhancing clinical explainability.
    }
    \label{fig:protomedx_architecture}
\end{figure*}

ProtoMedX introduces prototype-based deep learning for bone health assessment, enabling case-based reasoning and explanations that align with clinical decision-making. It learns class-representative prototypes in a unified feature space, making predictions by comparing inputs to these prototypes. Fig.~\ref{fig:protomedx_architecture} illustrates the overall architecture of the proposed framework.

\subsection{Feature Extraction and Encoding}
\textbf{Vision Branch:} We employ a frozen CrossViT backbone \cite{Chen2021} pre-trained on ImageNet for DEXA feature extraction, as it achieved superior standalone performance (See table \ref{tab:analysis}). CrossViT's dual-branch design captures both fine-grained trabecular patterns and global spinal morphology, critical for osteoporosis assessment. The frozen backbone outputs $\mathbf{f}^{\text{img}}\in\mathbb{R}^{1151}$, which is fed to a projection MLP $(1151\!\rightarrow\!512\!\rightarrow\!256)$ with BatchNorm and $0.3$ dropout, producing
\[
\mathbf{h}_i=\phi_i(\mathbf{f}^{\text{img}})\in\mathbb{R}^{256}.
\]

\noindent\textbf{Tabular Branch:} We encode the clinical features as a vector $\mathbf{x}^{\text{tab}}\in\mathbb{R}^{11}$ (Age, sex, weight, height, previous fracture, parent fracture hip, currently smoking, glucocorticoids, rheumatoid arthritis, secondary osteoporosis, and alcohol intake of 3 or more units/day) \textit{via}
\[
\mathbf{h}_t = \phi_t(\mathbf{x}^{\text{tab}}),\qquad
\phi_t:\mathbb{R}^{11}\!\rightarrow\!\mathbb{R}^{64},
\]
where $\phi_t$ is a two-layer MLP with residual connections and dropout.

\subsection{Cross-Modal Fusion}
The encoded image and tabular embeddings are first combined by a cross-modal attention block, producing a joint representation:
\[
\mathbf{h}_{\mathrm{fused}}
      = \operatorname{CrossAtt}\bigl(\mathbf{h}_i,\mathbf{h}_t\bigr)
      \in\mathbb{R}^{256}.
\]
\vspace{4pt}

\noindent
During prototype matching, we apply an adaptive, input-dependent weighting of modality-specific similarity scores:
\[
\alpha = \sigma\!\bigl(g([\mathbf{h}_i;\mathbf{h}_t])\bigr),\qquad
\mathrm{sim}(\mathbf{z})
      = \alpha\,S_{\text{img}}(\mathbf{z})
      + (1-\alpha)\,S_{\text{tab}}(\mathbf{z}),
\]
where $g$ is a linear layer and $S_{\text{img}},\,S_{\text{tab}}$ are cosine similarities in the image and tabular prototype subspaces, respectively. This gating strategy improved top-1 accuracy by $1.08\%$ compared with uniform weighting. The fused vector $\mathbf{h}_{\mathrm{fused}}$ is subsequently fed to the prototype layer for classification.

\subsection{Prototype Learning}
ProtoMedX learns $K=6$ prototypes per class, where each prototype represents a typical patient case. This value was determined through empirical validation: $K<6$ failed to capture intra-class diversity (e.g., early vs. advanced osteoporosis), while $K>6$ introduced redundant prototypes without improving classification. We employ a dual prototype architecture, maintaining separate spaces for visual ($\mathbf{p}_{c,k}^{img} \in \mathbb{R}^{128}$) and tabular ($\mathbf{p}_{c,k}^{tab} \in \mathbb{R}^{64}$) features while performing classification in the fused space. This enables modality-specific interpretation. Ablation shows that removing dual prototypes reduces accuracy by 4.33\%.

Following contrastive learning principles \cite{chen2020simple}, we optimise prototypes through a composite loss:
\begin{equation}
\mathcal{L}_{proto} = \mathcal{L}_{class} + \lambda_{sep}\mathcal{L}_{sep} + \lambda_{ctr}\mathcal{L}_{center}
\end{equation}

where $\mathcal{L}_{class}$ ensures prototypes alone can classify \textit{via} softmax over cosine similarities, $\mathcal{L}_{sep}$ maintains inter-class margins using a triplet-like loss \cite{zheng2021semi}, and $\mathcal{L}_{center}$ compacts intra-class distributions \cite{wen2016centerloss}. This creates well-separated clusters (Fig.~\ref{fig:tsne}) with osteoporosis prototypes clustering at low T-scores (-3.48 average).

To maintain explainability, prototypes are periodically projected onto the nearest training examples, ensuring that each represents an actual patient case rather than abstract features.

\subsection{Multi-Task Learning}

Traditional approaches inadequately exploit the fact that the bone density has a continuous nature and have notable limitations. Tabular methods heavily rely on external BMD values, achieving only 59\% accuracy without them (~Table\ref{tab:analysis}(b)). Two-stage vision frameworks that first predict BMD to make the diagnosis suffer from classification errors due to sharp clinical thresholds at T=-1 and T=-2.5 (See Table~~\ref{tab:analysis}(c)). Direct classification ignores continuous bone-density variations, plateauing at about 76\% accuracy, although not far from the standard cross-entropy approaches (See Table~\ref{tab:analysis}(a)).

We address these issues using multi-task learning, jointly optimising for classification and T-score regression. This encourages the model to learn continuous bone-density representations, smooth decision boundaries, and regularise training. Formally:
\begin{equation}
\mathcal{L}_{ProtoMedX} = \mathcal{L}_{cls} + \lambda_1\mathcal{L}_{reg} + \lambda_2\mathcal{L}_{proto}
\end{equation}
where $\mathcal{L}{cls}$ handles classification, $\mathcal{L}{reg}$ predicts T-scores, and $\mathcal{L}_{proto}$ is the prototype learning loss. Including regression ($\lambda_1=0.3$) significantly improves accuracy, removing it results in a 2.46\% accuracy drop. Thus, explicitly modelling bone density continuity substantially enhances clinical predictions which, in the end, are in a form of 3-class classification (Normal/ Osteopenia/Osteoporosis). The $\lambda$ multipliers  were determined after grid search.

\subsection{Clinical Explanations}
ProtoMedX provides clinically interpretable, multi-level explanations for each diagnosis. (See Sec.\ref{subsec:Clinical_Explanations} and Fig.~\ref{fig:protomedx_horizontal}).

\noindent\textbf{Classification Confidence:} Model confidence is derived from the weighted k-NN voting among prototypes:
\begin{equation}
\mathcal{C}=\frac{\sum_{i\in\mathcal{N}_c}e^{-d_i/\tau}}
{\sum_{j=1}^k   e^{-d_j/\tau}},
\end{equation}
where $\mathcal{N}_c\subseteq\{1,\dots,k\}$ are the $k$ nearest prototypes of the predicted class $c$, $d_i$ is the Cosine distance to prototype $i$, and $\tau$ controls the sharpness of the weighting.

High confidence (e.g. \( >90\% \)) indicates strong consensus; low confidence (e.g. \( <60\% \)) highlights borderline or uncertain cases. An advantage of the proposed ProtoMedX is the ability to analyse the decisions.

\noindent\textbf{Prototype-based Reasoning:} Each decision is supported by the most similar $k$ ($k=3$) prototypes, displaying source patient ID, clinical features, influence weight, and prototype representativeness. 

\noindent\textbf{Feature-Level Analysis:} For each clinical feature $j$, we compute relative deviation from class norms  (for class c):
\begin{equation}
\delta_j = \frac{|x_j^{tab} - \mu_{c,j}|}{\max(\mu_{c,j}, 1)}
\end{equation}
Features with $\delta_j > 0.5$ signal clinically atypical values.

\noindent\textbf{Voting Visualisation:} Bar plots show vote distributions, helping identify borderline versus confident classifications.

\noindent\textbf{Misclassification Analysis:} ProtoMedX provides confidence, which is an indication of a possible misclassification as described earlier. Furthermore, for the errors, in real-world settings, confidence scores and the explanations should aid the analysis. For our study, we visualized both predicted and true class prototypes, supporting error analysis. Most mistakes occur near diagnostic boundaries (e.g., Normal→Osteopenia). (See Fig. \ref{fig:protomedx_horizontal} for an example).

\subsection{Prototype-Based Classification}
During inference, ProtoMedX performs explainable classification through k-nearest neighbour search:

\begin{algorithm}[H]
\caption{ProtoMedX Inference and Explanation Generation}
\begin{algorithmic}[1]
\State \textbf{Input:} Patient $(\mathbf{x}^{img}, \mathbf{x}^{tab})$, Learned prototypes $\mathbf{P}$
\State \textbf{// Feature extraction and fusion}
\State $\mathbf{z}, \alpha \leftarrow \text{EncodeFuse}(\mathbf{x}^{img}, \mathbf{x}^{tab})$ 
\State \textbf{// k-NN classification}
\State Compute distances: $d_{c,k} = \|\mathbf{z} - \mathbf{p}_{c,k}\|_2$ for all prototypes
\State Select k=3 nearest prototypes: $\mathcal{N} = \{(c_i, k_i, d_i)\}_{i=1}^3$
\State Calculate weights: $w_i = \exp(-d_i/\tau)$, normalize to sum to 1
\State Vote: $v_c = \sum_{i: c_i=c} w_i$ for each class $c$
\State Predict: $\hat{y} = \arg\max_c v_c$
\State \textbf{// Generate clinical explanation}
\State Confidence: $\mathcal{C} = v_{\hat{y}}$ (winning vote weight)
\State Retrieve prototype metadata: patient IDs, T-scores, clinical features
\State Compute feature deviations: $\delta_j = |x_j^{tab} - \mu_{\hat{y},j}|/\max(\mu_{\hat{y},j}, 1)$
\State Flag atypical features where $\delta_j > 0.5$
\State \textbf{Return:} Class $\hat{y}$, confidence $\mathcal{C}$, similar patients $\mathcal{N}$, 
\State \quad\quad\quad modality weight $\alpha$, feature analysis $\{\delta_j\}$
\end{algorithmic}
\end{algorithm}

This approach achieves 89.8\% accuracy with k-NN classification, comparable to the less explainable approach of neural network classification layers instead of K-NN (89.6\%).

\section{Experiments}
\subsection{Dataset and Preprocessing}
\label{subsec:dataset}
We utilise a clinical dataset of 4,160 lumbar spine DEXA scans with corresponding patient records. Each scan includes DEXA produced T-scores and clinical features (age, sex, height, weight, previous fractures, parent fractured hip, current smoker, glucocorticoids, rheumatoid arthritis, secondary osteoporosis, and alcohol intake).

Images undergo standardised preprocessing: adaptive histogram equalisation for contrast enhancement and Gaussian filtering ($\sigma=1.0$) for noise reduction. All images are resized to 240×240 for CrossViT feature extraction. The dataset exhibits class imbalance typical of clinical populations: Normal (45\%), Osteopenia (38\%), and Osteoporosis (17\%).

\subsection{Training Configuration}
We use $80\%$ of the data for training (with an internal $10\%$ validation split) and $20\%$ as a held-out test set for evaluation. Differential learning rates are set for each component: $5 \times 10^{-5}$ for frozen image features, $5 \times 10^{-4}$ for tabular layers trained from scratch, and $1 \times 10^{-3}$ for prototype adaptation.

Training uses the AdamW optimiser with weight decay $10^{-4}$, cosine annealing schedule, and early stopping (patience=15). Prototypes are initialised via k-means clustering on training features and periodically projected to the nearest training examples to maintain explainability.

\subsection{Evaluation Protocol}
We evaluate using: (1) overall accuracy, (2) per-class precision/recall/F1, and (3) clinical agreement (predicted vs. true diagnostic category from T-scores). For ablations, we systematically remove components and measure performance degradation.

\subsection{Baseline Comparisons}
We compare our method against SOTA tabular approaches, vision-only models, other prototype models (see Table \ref{tab:analysis}), and recent SOTA bone health classification methods (see Table~\ref{tab:international}). For a fair comparison, we evaluated baselines and comparison in our aforementioned dataset and training splits, keeping the exact architecture, pre-processing and other architecture contributions claimed by each of the surveyed methods.
\section{Results}
\subsection{Quantitative Performance}

ProtoMedX demonstrates strong clinical alignment: 83.9\% diagnostic agreement, and 91.2\% sensitivity for Normal vs. Abnormal detection and SOTA performance with 89.2\% accuracy on three-class bone health classification.

 Detailed metrics show balanced performance: 89.3\% precision, 89.2\% recall, and 89.1\% F1-score. Per-class analysis reveals excellent Normal detection (92.1\% precision, 95.0\% recall), balanced Osteopenia performance (88.7\% precision, 78.6\% recall), and high Osteoporosis sensitivity (79.4\% precision, 92.6\% recall). Crucial for identifying severe cases requiring immediate intervention.

\begin{table}[h]
\centering
\caption{Bone Health Classification Performance  comparison across methods and modalities (\%)}
\label{tab:performance}
\resizebox{\columnwidth}{!}{%
\begin{tabular}{lccccc}
\toprule
\textbf{Model} & \textbf{Modality} & \textbf{Acc.} & \textbf{Prec.} & \textbf{Rec.} & \textbf{F1} \\
\midrule
\textbf{ProtoMedX} & Multi & \textbf{89.8} & \textbf{89.3} & \textbf{89.2} & \textbf{89.1} \\
MLP-MultiTask & Multi & 87.8 & 84.7 & 85.4 & 85.0 \\
MLP-SingleTask & Multi & 82.7 & 82.1 & 75.0 & 77.6 \\
CrossViT & Multi & 76.2 & 75.2 & 64.0 & 68.2 \\
MLP & Multi & 72.3 & 65.3 & 66.0 & 65.6 \\
\midrule
\textbf{MLP-MultiTask} & \textbf{Vision} & \textbf{87.8} & \textbf{85.1} & \textbf{85.4} & \textbf{85.2} \\
ProtoMedX-V & Vision & 87.5 & 83.6 & 85.3 & 84.2 \\
MLP-SingleTask & Vision & 82.3 & 82.1 & 73.3 & 76.1 \\
CrossViT & Vision & 74.7 & 69.5 & 67.7 & 68.5 \\
XGBoost & Vision & 73.2 & 65.7 & 66.0 & 65.8 \\
MLP & Vision & 73.2 & 65.8 & 66.1 & 65.9 \\
\bottomrule
\end{tabular}%
}
\label{tab:full-experiments}
\end{table}

Note that MLP-MultiTask jointly estimates T-scores and bone health classification, whereas MLP-SingleTask estimates T-scores only, which are then mapped to the classification boundaries. The multi-task approach consistently improves performance across modalities.

\subsection{Ablation Study}
Component-wise ablation reveals each architectural contribution:

\begin{table}[h]
\centering
\caption{Ablation study of ProtoMedX components}
\label{tab:ablation}
\begin{tabular}{lcc}
\toprule
Configuration & Accuracy (\%) & $\Delta$ \\
\midrule
Full ProtoMedX & \textbf{89.8} & - \\
\quad w/o gate mechanism & 88.72 & -1.08 \\
\quad w/o multi-task & 87.34 & -2.46 \\
\quad w/o cross-attention & 85.71 & -4.09 \\
\quad w/o prototypes & 85.47 & -4.33 \\
Baseline (no components) & 84.99 & -4.81 \\
\bottomrule
\end{tabular}
\end{table}

Prototype learning and cross-modal attention provide the largest gains (4.33\% and 4.09\%), validating our architectural choices. Multi-task learning adds 2.46\%, confirming that continuous T-score supervision, alongside categorical classification, enhances overall performance.

\begin{table}[h]
\centering
\caption{Empirical‰ demonstration of key architectural choices}
\label{tab:analysis}
\begin{tabular}{@{}l@{\hspace{8pt}}c@{\hspace{8pt}}c@{}}
\toprule
\multicolumn{3}{c}{\textbf{(a) Vision Backbone Comparison}} \\
\midrule
\textbf{Backbone} & \textbf{Acc (\%)} & \\
\midrule
ResNet/VGG & 67/73 & \\
DenseNet/ViT & 71/74 & \\
\textbf{CrossViT~\cite{Chen2021}} & \textbf{76} & \\
\midrule
\multicolumn{3}{c}{\textbf{(b) Tabular Methods: BMD Dependency}} \\
\midrule
\textbf{Method} & \textbf{w/ BMD} & \textbf{w/o BMD} \\
\midrule
XGBoost/GB & 85 & 53/54 \\
Decision Tree & 83 & \textbf{59} \\
Logistic Reg. & 79 & 55 \\
\midrule
\multicolumn{3}{c}{\textbf{(c) BMD Regression vs Classification}} \\
\midrule
\textbf{Model} & \textbf{MSE} & \textbf{Acc (\%)} \\
\midrule
ViT & \textbf{0.02} & 68.7 \\
DenseNet & 0.03 & 46.9 \\
CrossViT & 0.05 & \textbf{71.4} \\
\bottomrule
\end{tabular}
\end{table}

\paragraph{Validation of other architectural choices: }Table~\ref{tab:analysis}(a) validates CrossViT selection for vision backbone. Table~\ref{tab:analysis}(b) reveals tabular methods' critical dependency on BMD values, without them, accuracy drops from 85\% to between 53 and 59\%. This justifies our vision-centric approach, as clinical BMD measurements may be unavailable in settings where, for instance, X-Ray imagery is used instead of the more costly DEXA machines. Table~\ref{tab:analysis}(c) demonstrates that low BMD regression MSE does not guarantee high classification accuracy. Despite ViT achieving lowest MSE (0.02), CrossViT's higher MSE (0.05) yields superior classification ($71.4\%$ vs $68.7\%$), because small BMD errors near diagnostic thresholds ($T=-1, T=-2.5$) cause misclassifications. With regression-then-prediction tasks always showing inferior classification performance. This motivates our multitask approach that jointly optimizes both objectives.

\subsection{Prototype Explainability}
\begin{figure}[h]
    \centering
    \includegraphics[width=\columnwidth]{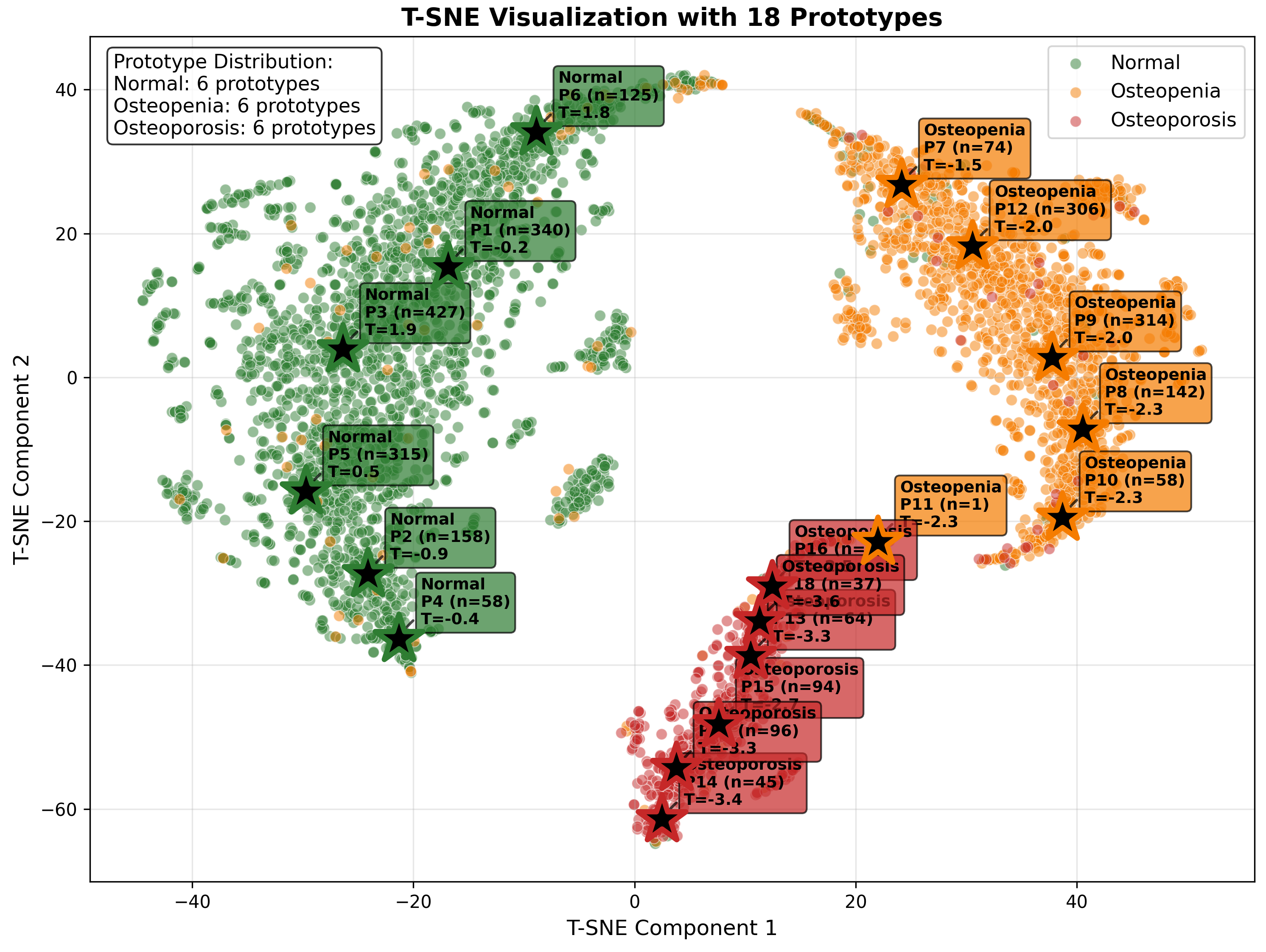}
    \caption{T-SNE analysis of Fused Prototype feature space with 18 learned prototypes showing clear class separation.}
    \label{fig:tsne}
\end{figure}

\begin{figure}[h]
    \centering
    \includegraphics[width=\columnwidth]{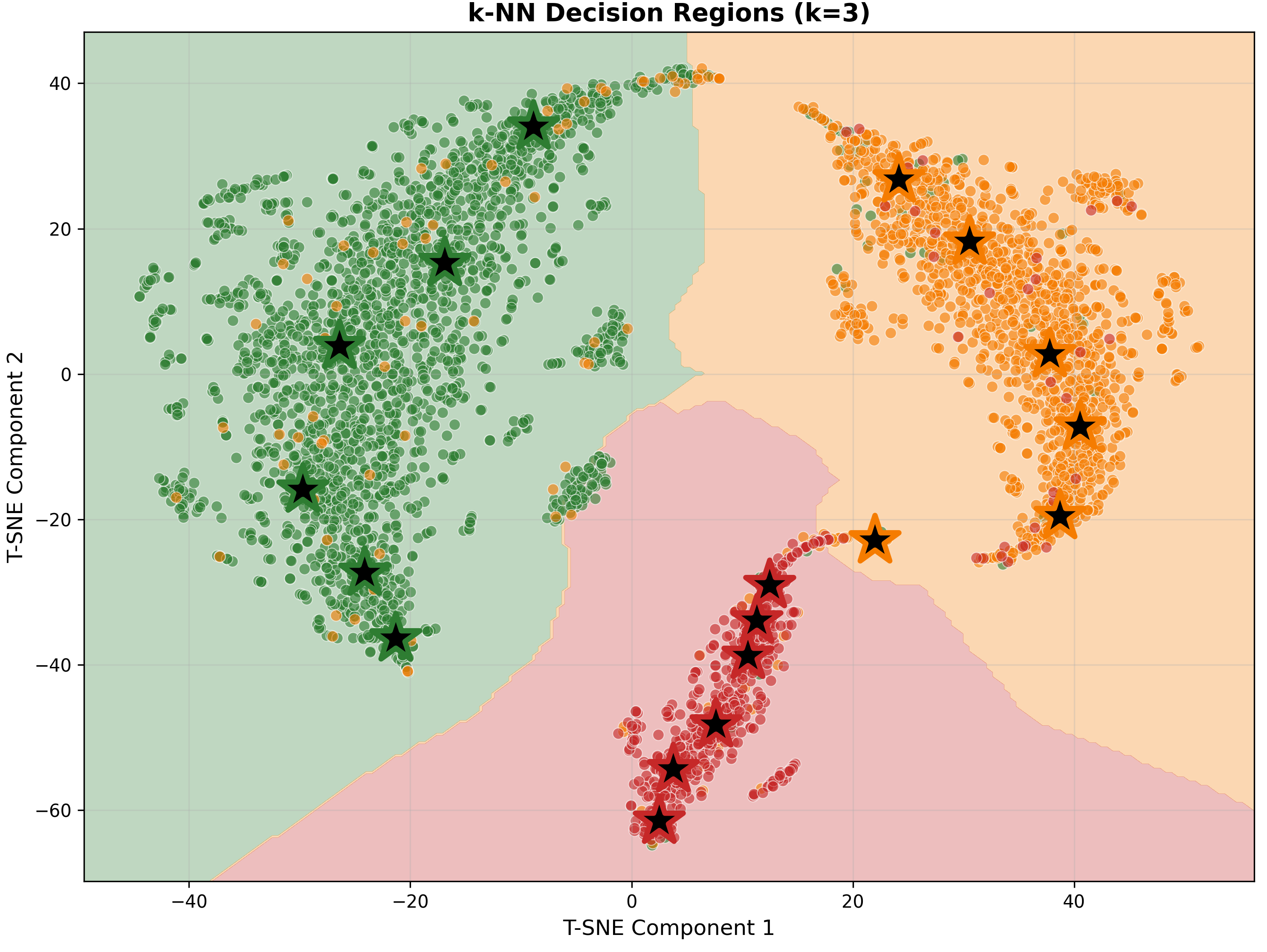}
    \caption{T-SNE analysis of k-NN decision boundaries ($k=3$) in Fused Prototype feature space demonstrating how prototypes define diagnostic regions.}
    \label{fig:tsne_knn_prototypes2}
\end{figure}

T-SNE visualization (Figure~\ref{fig:tsne}) reveals well-separated prototype clusters with clear decision boundaries (Figure~\ref{fig:tsne_knn_prototypes2}). Normal prototypes (green) cluster tightly at positive T-scores, while Osteoporosis prototypes (red) show broader distribution reflecting disease heterogeneity. The k-NN decision regions demonstrate how prototypes create explainable classification boundaries in the learned feature space (Figure~\ref{fig:tsne_knn_prototypes2}).

\subsection{Clinical Explanations} 
\label{subsec:Clinical_Explanations}

\begin{figure*}[t]
\centering
\includegraphics[width=\textwidth,clip,trim=0 5 0 5]{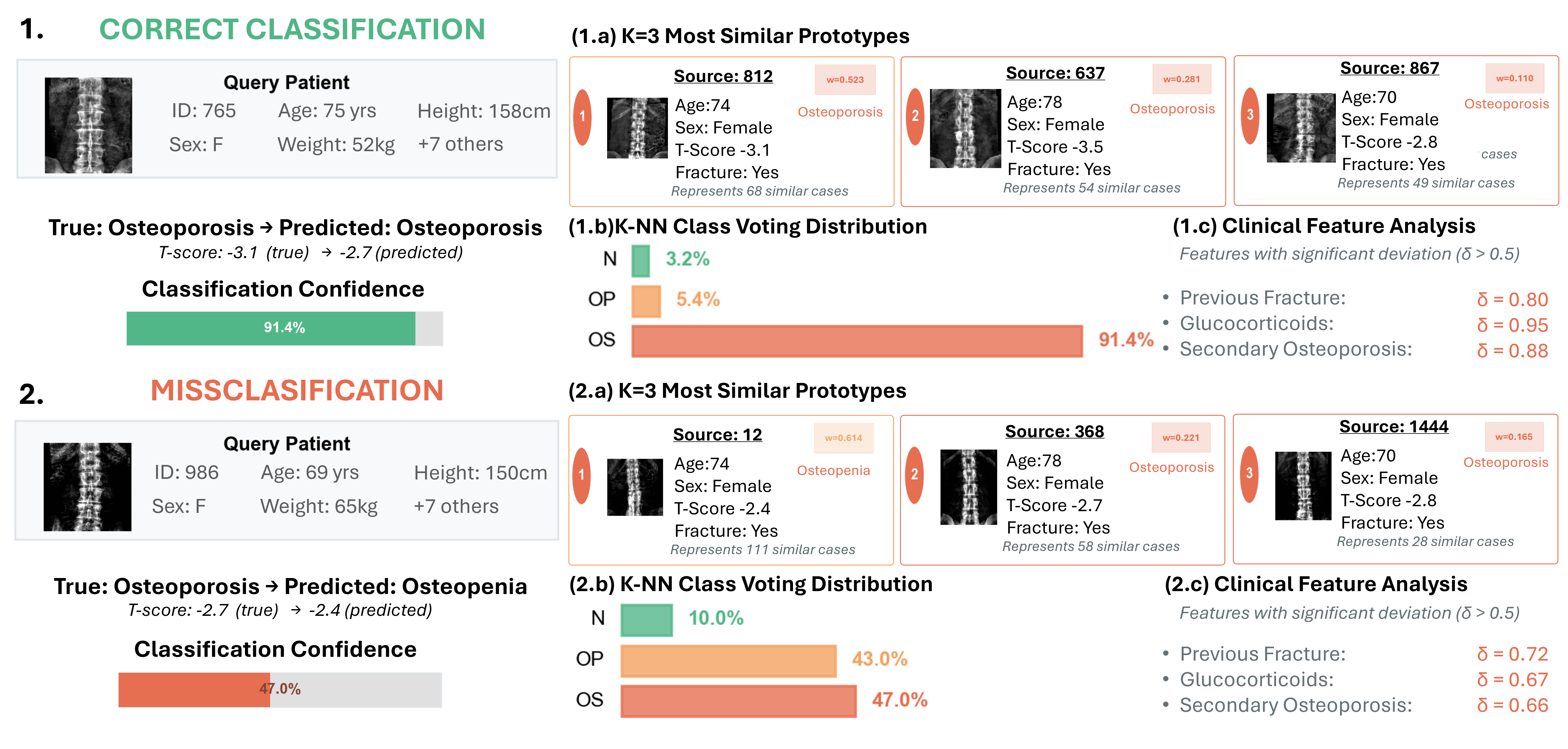}
\caption{ProtoMedX Clinical Explanations. (1) Correct classification and (2) misclassification examples. Each panel includes: (a) prototype similarity with annotated clinical metadata, (b) model confidence and class voting distribution, and (c) clinical feature deviation analysis.}
\label{fig:protomedx_horizontal}
\end{figure*}

Figure~\ref{fig:protomedx_horizontal} presents ProtoMedX's explanation interface, which traces each prediction to the k=3 most similar prototypes with their clinical characteristics and similarity weights.

Panel (1) shows a correct osteoporosis classification with 91.4\% confidence. All three prototypes belong to the osteoporosis class ($w= 0.523, 0.281, 0.110$), demonstrating strong consensus. The feature analysis identifies key risk factors: previous fracture ($\delta = 0.80$), glucocorticoid use ($\delta = 0.95$), and secondary osteoporosis ($\delta = 0.88$), providing clear clinical justification.

Panel (2) demonstrates the model's transparency in error analysis. ProtoMedX incorrectly predicts osteopenia instead of osteoporosis, but the low confidence (47.0\%) immediately flags uncertainty. The ambiguous voting distribution and mixed-class prototypes reveal the patient lies at a diagnostic boundary. Feature deviations ($\delta = 0.66$--0.72) are moderate, indicating weaker discriminative signals.

This framework enables clinicians to assess prediction reliability. Correct predictions average 85.3\% confidence versus 48.9\% for misclassifications. Most errors occur at diagnostic boundaries where patients exhibit characteristics of multiple classes, precisely where transparent decision support is most valuable.
\subsection{Comparison to Competitors}
\begin{table}[h]
\centering
\caption{Comparative analysis of Bone Health Classification approaches. } 
\label{tab:international}
\resizebox{\columnwidth}{!}{%
\begin{tabular}{llcc}
\toprule
\textbf{Institution} & \textbf{Model} & \textbf{Modality} & \textbf{Acc (\%)} \\
\midrule
\textbf{Ours} & \textbf{ProtoMedX} & \textbf{Multi} & \textbf{89.8} \\
\textbf{Ours} & \textbf{ProtoMedX-V} & \textbf{Vision} & \textbf{87.58} \\
Okayama \cite{sukegawa2022identification} & EfficientNet-b7 & Vision & 75.7 \\
Ours & ProtoPNet-MT & Vision & 72.4 \\
Ours & xDNN (CrossViT backbone) & Vision & 71.0 \\
Stanford \cite{luan2025application} & ResNet-50 & Vision & 69.2 \\
Toosi \cite{sarmadi2024comparative} & ViT & Vision & 68.3 \\
Anna \cite{9752533} & VGG16 & Vision & 66.4 \\
Ours & ProtoPNet & Vision & 64.1 \\
Anna \cite{9752533} & Inception v3 & Vision & 62.9 \\
\bottomrule
\end{tabular}%
}
\end{table}

ProtoMedX significantly outperforms existing approaches (Table~\ref{tab:international}), achieving 14 to 27\% absolute improvement. Notably, our vision-only variant surpasses all prior methods while providing inherent explainability.


\section{Conclusion} 
We presented ProtoMedX, the first prototype-based framework for explainable bone health classification. By rethinking machine learning approaches to bone health classification through case-based reasoning, ProtoMedX achieves $89.8\%$ accuracy, surpassing prior methods by $14$ to $27\%$, while offering inherent explainability through prototypes and decision boundaries that clinicians can interpret and critique to support their medical judgment.

Three innovations drive this performance: (1) Prototype-based reasoning which adds $4.33\%$, while enabling transparent explanations (2) Cross-modal attention which ensures balanced fusion, contributing $4.09\%$, and (3) Multi-task learning leveraging bone density continuity which improves accuracy by 2.46\% over single-task classification.

Unlike \textit{post hoc} interpretation methods, ProtoMedX is explainable-by-design through direct comparison to learned prototypes. Each prediction traces to similar prototypes with complete clinical profiles, enabling clinicians to understand the reasoning behind each classification. The dual prototype architecture separately models visual (trabecular thinning and porous bone) and clinical features (demographics, risk factors) before fusing them for diagnosis.

Analysing the limitations of the proposed architecture, we notice that the marginal improvement from multi-modal fusion (87.58\% to 89.8\%) suggests unexploited complementary information. Additionally, full-image prototypes lack the granular localisation needed for precise clinical guidance. Future work would, therefore, explore explainable-by-design attention-guided prototype localisation and longitudinal modelling for monitoring disease progression.

ProtoMedX demonstrates that it is possible to achieve improved explainability without sacrificing performance. By aligning with clinical reasoning we advance toward deployable xAI that enhances, rather than replaces, clinical judgment.
\section{Acknowledgments}

This study is approved by the North West- Preston Research Ethics Committee (REC ref 21/NW/0309). This work is supported by ELSA – European Lighthouse on Secure and Safe AI funded by the European Union under grant agreement No. 101070617.

{
    \small
    \bibliographystyle{ieeenat_fullname}
    \bibliography{main}
}


















\end{document}